\documentclass{article}

\usepackage[preprint]{neurips_2026}
\workshoptitle{Trustworthy AI for Good (AI4GOOD)}

\makeatletter
\renewcommand{\@notice}{}
\makeatother

\usepackage[utf8]{inputenc}
\usepackage[T1]{fontenc}
\usepackage{hyperref}
\usepackage{url}
\usepackage{booktabs}
\usepackage{amsfonts}
\usepackage{amsmath,amssymb}
\usepackage{nicefrac}
\usepackage{microtype}
\usepackage{xcolor}
\usepackage{graphicx}
\usepackage{placeins}
\usepackage{enumitem}
\usepackage{algorithm}
\usepackage{algorithmic}
\usepackage{tikz}
\usetikzlibrary{arrows.meta,positioning}
\usepackage{pgfplots}
\pgfplotsset{compat=1.16}

\newcommand{\gemmm}{GEM-MM}
\newcommand{\best}[1]{\textbf{#1}}
\newcommand{\secbest}[1]{\underline{#1}}
\newcommand{\metdown}{$\downarrow$}
\newcommand{\metup}{$\uparrow$}

\title{\gemmm{}: Reducing VLM Hallucination with Entropy-Guided Preference Alignment}

\author{%
  Yuanshan Chua \\
  College of Computing and Data Science (CCDS), Nanyang Technological University (NTU), Singapore \\
  \texttt{ychua060@e.ntu.edu.sg} \\
  \And
  Xuejiao Zhao\thanks{Corresponding author.} \\
  College of Computer Science, Sichuan University, China\\
  College of Computing and Data Science (CCDS), Nanyang Technological University (NTU), Singapore \\
  \texttt{xjzhao@ntu.edu.sg} \\
}

\begin{document}

\maketitle

\begin{abstract}
Vision-language models (VLMs) are routinely aligned with offline pairwise
objectives inherited from text-only RLHF, but multimodal preference data is scarce
and every response token is treated as equally informative.
Under a fixed, small preference budget this makes the choice of
\emph{pair construction} nearly irrelevant: standard DPO-style multimodal
objectives, image-conditional preference optimization, and self-improvement pair
mining all land within $1.1$ points of one another on held-out MM-RLHF preference
metrics and within $0.5$ points on generative hallucination rate, and the
fine-tuned (SFT) control that initializes them falls inside the same band.
We call this regime \textbf{entropy-blind preference optimization}: the learning
signal never looks at \emph{where} the policy is uncertain, even though in
chain-of-thought VLM responses uncertainty concentrates on a small set of
\emph{fork} tokens at which the model commits to a visual claim.
We introduce \gemmm{}, which extends entropy-guided preference modeling to VLMs:
it samples on-policy candidates, rewards high-entropy reasoning forks while
penalizing uncertain final answers, and updates the full model with a
group-normalized policy-gradient step.
On a locked 3000-prompt MM-RLHF budget with Qwen3-VL-4B-Instruct, \gemmm{}
attains the best entropy-depth preference score ($61.7\%$ vs.\ $57.3$--$58.4\%$
for the preference-tuned baselines) and the best objective-agnostic implicit-reward
preference accuracy ($41.4\%$), and most importantly reduces AMBER generative hallucination
from $23.4$--$24.0$ to \best{$20.6$} and CHAIR from $5.0$--$5.1$ to \best{$4.6$}
at matched caption coverage.
It is also the best system on all three HallusionBench granularities, while a
conservative operating point of the same objective raises the near-chosen rate to
$62.4\%$.
An off-the-shelf multimodal reward model that never saw our objective ranks
\gemmm{} first among the systems compared, and two such reward models independently
confirm our two main ablations at $p<0.001$.
We report per-metric confidence intervals and paired significance tests, state
explicitly which of our metrics share functional form with the training reward,
and document a collapse mode that appears when training is pushed past
the fair budget.
Code, configurations, and the evaluation harness are available at
\url{https://github.com/SNOWTEAM2023/GEM-MM}.
\end{abstract}

\section{Introduction}

Vision-language models (VLMs) now support competitive multimodal
reasoning~\citep{liu2023visual,liu2024improved,bai2025qwen25vl,qwen2025qwen3vl},
but remain prone to fluent responses that are not supported by the
image~\citep{li2023pope,rohrbach2018object}.
Preference alignment is the standard remedy, via a learned reward and
RL~\citep{ouyang2022training,christiano2017deep} or a closed-form pairwise
loss~\citep{rafailov2023direct}.
Multimodal work has largely inherited this recipe, improving pair
construction~\citep{sun2024aligning,li2023silkie,yu2025rlaifv,zhang2025mmrlhf,wang2025sima,zhou2024aligning}
or image conditioning~\citep{wang2024mdpo,huang2024vdpo}.

Under a few-thousand-pair budget, the choice of pairs barely moves the needle
(Table~\ref{tab:diagnostic}). Fixing initialization, prompts, template, and
token budget, a DPO-style multimodal objective, an image-conditional
objective, and a self-improvement pipeline are statistically indistinguishable
on held-out preference accuracy and plateau at the same AMBER hallucination rate.
These objectives consume a sequence-level likelihood ratio in which every
response token contributes symmetrically. We call this \textbf{entropy-blind
preference optimization}.

Token entropy in a chain-of-thought is concentrated on a minority of
\emph{fork} tokens which can be defined as the 'branch points' at which the model commits to a visual
claim. For example, final-answer entropy should be low if the question is
resolved~\citep{wang2025beyond8020,agarwal2025unreasonable,cui2025entropy,farquhar2024detecting}.
GEM~\citep{zhao2026gem} formalized this fork-versus-final asymmetry for text-only
alignment with a group-normalized policy gradient (SEGA), building on generative
few-shot reward inference~\citep{zhao2025gfriendgenerativefewshotreward,ZHAO2026116645}.
Complementary entropy-aware multimodal work includes personalized concept learning~\citep{bai2025online};
related deployment-oriented copilot systems appear in~\citep{zhao2025medrag,ijcai2025p1278}.
In this paper, we ask whether that signal still holds for VLMs, 
where image tokens precede text, splits are computed over decoded CoT, and the primary failure mode shifts to visual hallucination.

\paragraph{Contributions.}
(i)~We quantify entropy-blind saturation: three pair-based objectives on identical
data differ by at most $1.1$ preference points and $0.5$ AMBER Hal
(Table~\ref{tab:diagnostic}).
(ii)~We propose \gemmm{}, which scores on-policy VLM candidates with a
fork-versus-final entropy functional and updates all parameters with a
group-normalized policy gradient (Figure~\ref{fig:pipeline}).
(iii)~On a locked 3000-prompt MM-RLHF budget, \gemmm{} is best on entropy-depth
and implicit-reward preference, and it reduces AMBER hallucination to $20.6$ and
CHAIR to $4.6$ at matched coverage.
(iv)~To keep our evaluation honest, we treat the training-form metric strictly as a diagnostic and rely on objective-agnostic CIs and paired tests. 
Along the way, we also uncover a length/format confound and a collapse mode during continued training (Section~\ref{sec:analysis}).

\section{Preliminaries}
\label{sec:prelim}

\paragraph{Multimodal preference data.}
A multimodal preference dataset is a set of tuples
$\mathcal{D}=\{(x_I,x_Q,y^{+},y^{-})\}$, where $x_I$ is an image, $x_Q$ a question
or instruction, and $y^{+}$ is annotated as preferred over $y^{-}$.
We write $x=(x_I,x_Q)$ and let $\pi_\theta(y\mid x)$ be a VLM policy that consumes
the image and text tokens.

\paragraph{Reward-model and direct objectives.}
Classical RLHF assumes preferences follow a Bradley--Terry
model~\citep{bradley1952rank}, fits a reward $r_\phi$ by maximum likelihood, and
then optimizes $\pi_\theta$ against $r_\phi$ with a KL anchor to a reference
policy $\pi_{\mathrm{ref}}$, typically by
PPO~\citep{schulman2017proximal,ouyang2022training}.
DPO~\citep{rafailov2023direct} removes the explicit reward by reparameterizing the
KL-regularized optimum as a pairwise logistic loss on the \emph{implicit reward}
$\Delta_\theta(y)=\log\pi_\theta(y\mid x)-\log\pi_{\mathrm{ref}}(y\mid x)$.
A family of variants alters the link function or the normalization, including
IPO~\citep{gheshlaghi2024general}, KTO~\citep{ethayarajh2024model},
SimPO~\citep{meng2024simpo}, ORPO~\citep{hong2024orpo}, and
ranking losses~\citep{yuan2023rrhf,zhao2023slic,song2024pro,liu2025lipo}. These methods share a key limitation: 
a sequence-level likelihood ratio that treats all response tokens the same.

Multimodal variants add image conditioning to the same skeleton: MM-DPO, as
released with MM-RLHF~\citep{zhang2025mmrlhf}, reweights the pairwise term by an
annotation-quality factor, and mDPO~\citep{wang2024mdpo} adds a conditional
preference term over perturbed images plus an anchor that keeps
$\log\pi_\theta(y^{+})$ from drifting downward.

\paragraph{On-policy alternatives.}
Instead of consuming fixed pairs, one may sample candidates from the current
policy and reweight them by a scalar reward, as in
REINFORCE-style~\citep{williams1992simple,ahmadian2024back} and group-normalized
objectives~\citep{shao2024deepseekmath,deepseek2025r1}.
\gemmm{} lives in this family: it never consumes $y^{+}$ or $y^{-}$ as targets
during optimization; the preference pairs are used only to define the prompt pool
and the held-out evaluation. This distinction matters when interpreting our
results, and we return to it in Section~\ref{sec:setup}.

\paragraph{Notation for entropy.}
For a decoded response $y=(y_1,\dots,y_T)$ with next-token logits $\ell_t$ we write
$p_t=\mathrm{softmax}(\ell_t)$ and
\begin{equation}
\label{eq:entropy}
H_t \;=\; -\!\sum_{v\in\mathcal{V}} p_t(v)\log p_t(v),
\end{equation}
so $H\in\mathbb{R}^{T}$ is the per-token entropy profile of the response under the
model being scored.

\section{Entropy-blind preference optimization}
\label{sec:diagnostic}

We fix the backbone (Qwen3-VL-4B-Instruct), the fine-tuned (SFT) initialization, the
prompt template, the decoding budget, and the exact set of 3000 MM-RLHF training
prompts, and vary only the alignment objective:
(i)~MM-DPO, a quality-weighted multimodal DPO~\citep{zhang2025mmrlhf};
(ii)~mDPO, which adds image-conditional preference and an anchor
term~\citep{wang2024mdpo};
(iii)~SIMA self-improvement, in which the pairs themselves are generated and
ranked by the model rather than annotated~\citep{wang2025sima}.
These three differ in what a pair is and in how the likelihood ratio is
conditioned, which is why these baselines were selected.

Table~\ref{tab:diagnostic} reports the outcome.

\begin{table}[!ht]
\caption{The saturation that motivates \gemmm{}. Identical initialization, prompts
(3000 MM-RLHF IDs), template, and token budget; only the alignment objective
differs. Spread = max $-$ min over the three preference-tuned rows. All three
objectives land inside one measurement interval of each other.}
\label{tab:diagnostic}
\centering
\setlength{\tabcolsep}{3.2pt}
\small
\begin{tabular}{lcccc}
\toprule
 & Depth & Implicit & A/B & AMBER \\
Objective & Pref\,\% & rew.\,\% & \% & Hal\,\metdown \\
\midrule
SFT only (no pairs) & 57.9 & 40.9 & 66.4 & 24.0 \\
\midrule
MM-DPO (quality-weighted) & 58.1 & 41.0 & 66.4 & 23.4 \\
mDPO (image-conditional) & 58.4 & 40.2 & 66.1 & 23.4 \\
SIMA (self-generated) & 57.3 & 40.8 & 66.0 & 23.9 \\
\midrule
\emph{spread (3 obj.)} & \emph{1.1} & \emph{0.8} & \emph{0.4} & \emph{0.5} \\
\midrule
\gemmm{} (entropy-guided) & \best{61.7} & \best{41.4} & \best{68.3} & \best{20.6} \\
\bottomrule
\end{tabular}
\end{table}
\FloatBarrier

\section{Method: \gemmm{}}
\label{sec:method}

\begin{figure}[t]
\centering
\includegraphics[width=\linewidth]{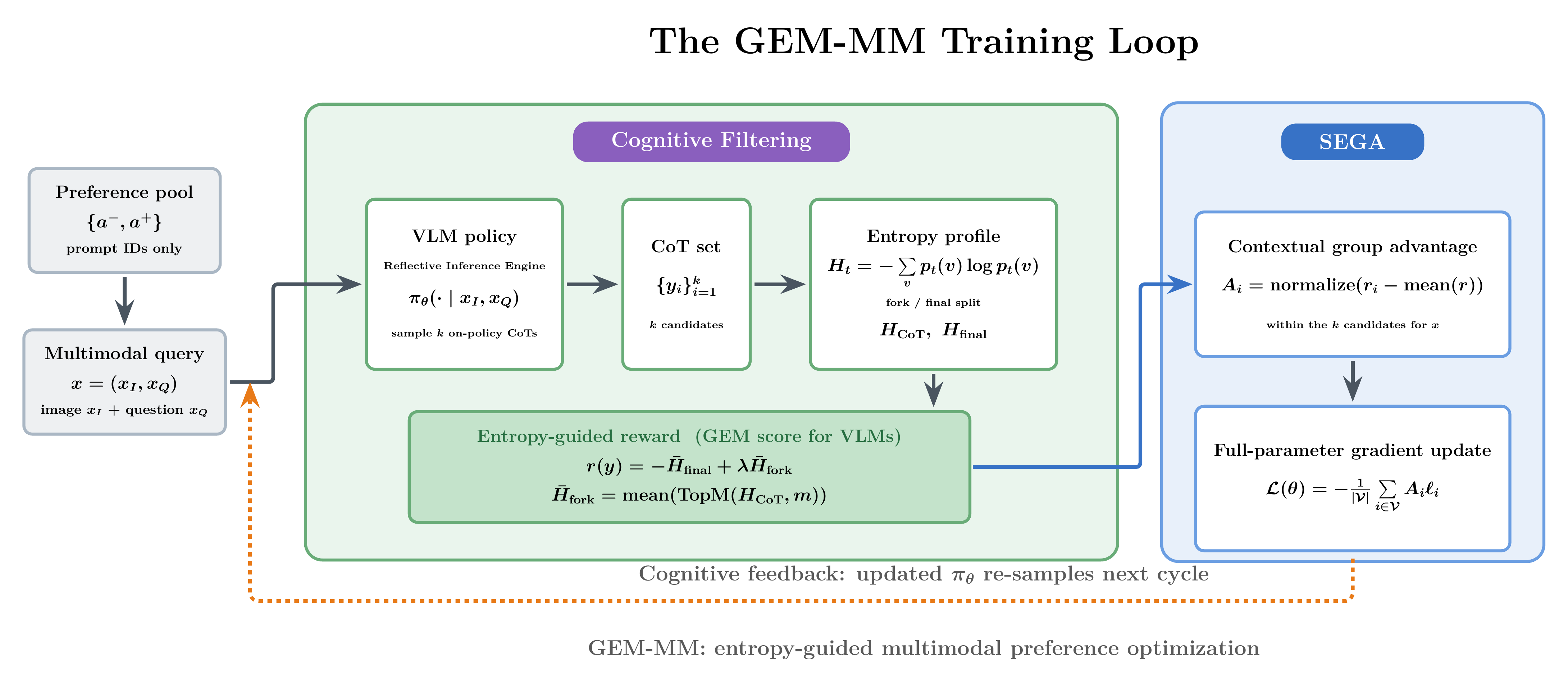}
\caption{Overview of the GEM pipeline instantiated by \gemmm{} for
vision-language models (adapted from~\citet{zhao2026gem}).
Given a multimodal query $x=(x_I,x_Q)$ (text-only $q$ in the original figure)
drawn from a preference pool $\{a^{-},a^{+}\}$, the Reflective Inference Engine
$\pi_\theta$ elaborates $k$ chain-of-thought candidates; Cognitive Filtering
computes per-token entropy and an entropy-guided score that rewards fork tokens
while penalizing uncertain final answers; SEGA forms a contextual group advantage
and applies a full-parameter gradient update.
The cognitive feedback loop returns \emph{only} to the inference engine
(updated policy), which then re-samples candidates on the next cycle and does
not re-enter the entropy-scoring block directly.}
\label{fig:pipeline}
\end{figure}

\gemmm{} keeps the prompt pool of a preference dataset but replaces pairwise
supervision with an on-policy loop over three stages: sample, score by entropy
structure, and update with group-normalized advantages
(Figure~\ref{fig:pipeline}).
The figure is the GEM closed-loop pipeline~\citep{zhao2026gem} that
\gemmm{} instantiates for VLMs.

\subsection{Candidate sampling}
For each prompt we draw $k$ completions
$y_1,\dots,y_k\sim\pi_\theta(\cdot\mid x_I,x_Q)$ with temperature $\tau$ and a
budget of $T_{\max}$ new tokens, using the same chat template and system prompt
used at evaluation time. Because sampling is on-policy, the reward is evaluated on
text the model actually produces, and the entropy profile in
Eq.~\eqref{eq:entropy} is the profile of the policy being updated.

\subsection{Fork/final split of the entropy profile}
A CoT response has two functionally different parts: a reasoning prefix, where
being uncertain is informative because the model is still considering
alternatives, and a final-answer suffix, where uncertainty indicates an unresolved
commitment. We therefore split each CoT answer as follows:
\begin{equation}
\label{eq:split}
s=\begin{cases}
\lfloor 0.75\,T\rfloor, & \text{if the decode contains ``Final Answer''},\\[2pt]
\lfloor 0.80\,T\rfloor, & \text{otherwise},
\end{cases}
\end{equation}
giving $H_{\mathrm{CoT}}=H_{1:s}$ and $H_{\mathrm{final}}=H_{s+1:T}$.
We deliberately choose a fixed fractional split over a marker-based one
to prevent the policy from trivially hacking the reward by moving the split point.

\subsection{Fork-entropy reward}
Rewarding \emph{mean} CoT entropy would incentivize uniform uncertainty everywhere,
which is the opposite of our goal.
Following the high-entropy-minority view of token
importance~\citep{wang2025beyond8020}, we keep only the top-$m$ entropy tokens of
the reasoning span, with $m=\max\!\left(1,\lfloor |H_{\mathrm{CoT}}|\cdot\rho\rfloor\right)$
for a top-$m$ ratio $\rho$:
\begin{align}
\label{eq:fork}
\bar H_{\mathrm{fork}}&=\mathrm{mean}\!\big(\mathrm{TopM}(H_{\mathrm{CoT}},m)\big),\\
\label{eq:final}
\bar H_{\mathrm{final}}&=\mathrm{mean}\!\big(H_{\mathrm{final}}\big),
\end{align}
where $\bar H_{\mathrm{final}}$ defaults to a large constant ($5.0$ nats) when the
final span is empty, so that degenerate decodes without an answer are penalized
rather than rewarded. The reward is
\begin{equation}
\label{eq:reward}
r(y)\;=\;-\,\bar H_{\mathrm{final}}\;+\;\lambda\,\bar H_{\mathrm{fork}},
\end{equation}
with $\lambda>0$, trading exploration at forks against decisiveness at the answer.
Eq.~\eqref{eq:reward} is the multimodal instantiation of the GEM
score~\citep{zhao2026gem}; the ablations in Section~\ref{sec:ablations} replace it
by its final-only term, its fork-only term, or a binary exact-match reward.

\subsection{Group-normalized update (SEGA)}
Within the $k$-candidate group for one prompt we form the baseline
$b=\frac1k\sum_i r_i$ and centered advantage $\tilde A_i=r_i-b$, then standardize
when the group has non-degenerate spread:
\begin{equation}
\label{eq:adv}
A_i=\frac{\tilde A_i-\mathrm{mean}(\tilde A)}{\mathrm{std}(\tilde A)+10^{-8}}
\quad\text{if } \mathrm{std}(\tilde A)>10^{-6},
\end{equation}
and $A_i=\tilde A_i$ otherwise. With the sequence log-likelihood
$\ell_i=\sum_t\log\pi_\theta(y_{i,t}\mid x,y_{i,<t})$ the objective is
\begin{equation}
\label{eq:sega}
\mathcal{L}(\theta)\;=\;\frac{1}{|\mathcal{V}|}\sum_{i\in\mathcal{V}}\big(-A_i\,\ell_i\big),
\end{equation}
optimized with AdamW over all parameters; $\mathcal{V}$ is the set of valid
(non-empty) candidates.

\paragraph{What the update does.}
The gradient of Eq.~\eqref{eq:sega} is
$-\frac{1}{|\mathcal{V}|}\sum_i A_i\nabla_\theta\ell_i$: candidates whose
entropy profile is more fork-heavy and more answer-decisive than their siblings on
the \emph{same} prompt get upweighted, and their siblings get downweighted, with
the group standardization in Eq.~\eqref{eq:adv} making the step size invariant to
the absolute scale of $r$ on that prompt.
Two consequences are worth stating because they are visible in our results.
First, because normalization is within-prompt, prompts on which all candidates
agree contribute almost nothing, so the effective data is the subset of prompts
where the policy is internally inconsistent, which is why the method is usable at
a 3000-prompt scale.
Second, because the reward is a function of the policy's own entropy, the loss can
be reduced either by making genuinely better-grounded decisions at fork tokens or
by reshaping the entropy profile stylistically. Ablating the fork term isolates the
first mechanism (Section~\ref{sec:ablations}), and we quantify the second as a
length/format bias in Section~\ref{sec:analysis}.

\subsection{Vision-language implementation}
\label{sec:impl}
The policy is Qwen3-VL-4B-Instruct~\citep{qwen2025qwen3vl}, initialized from a
multimodal SFT checkpoint trained on the same corpus.
We cap training generation at $T_{\max}{=}128$ tokens and evaluation generation
at $256$ tokens.
All reported \gemmm{} runs are full fine-tunes with gradient
checkpointing on a single 40\,GB A100.
The primary \gemmm{} model uses $\lambda{=}2.0$, $\rho{=}0.05$, $k{=}3$,
$\tau{=}0.9$, $\eta{=}1{\times}10^{-5}$, $T_{\max}{=}128$, one pass over 3000
prompts (3000 update cycles) and the fractional entropy split in Eq.~\eqref{eq:split}. 
Rather than converging to a single optimal setting, 
we show in Section~\ref{sec:analysis} that $\lambda$ and $\rho$ trace out an interpretable trade-off frontier.

\section{Experimental setup}
\label{sec:setup}

\paragraph{Data and budget.}
MM-RLHF~\citep{zhang2025mmrlhf} ships several preference modalities (short, long,
MCQ, safety, and video).
We restrict training and held-out preference evaluation to the \texttt{short}
image modality only: the first 3000 unique short IDs form the training pool for
\emph{every} method, and IDs 3001--4000 are held out and never trained on.

\paragraph{Systems compared.}
\emph{Base}: the instruction-tuned backbone, no alignment.
\emph{SFT}: supervised fine-tuning on the same corpus, which also initializes every
preference method.
\emph{MM-DPO}: multimodal DPO with dynamic per-sample scaling of $\beta$ by an
annotation-derived score factor, as released with
MM-RLHF~\citep{zhang2025mmrlhf} (scaling weight $0.1$).
\emph{mDPO}: image-conditional preference optimization~\citep{wang2024mdpo}
(image term weight $1.0$).
\emph{SIMA}: self-improvement pair mining~\citep{wang2025sima}; we generate
candidates from the SFT policy and rank them with a visual/format critic, then
train with the same pairwise trainer.

\paragraph{Evaluation protocol.}
\label{par:eval-depth}
We separate metrics by independence from the training signal.
Entropy-depth preference scores held-out pairs with Eq.~\eqref{eq:reward}
($r(y^{+})>r(y^{-})$); it shares functional form and $(\lambda,\rho)$ with the
training reward, so we report it as a depth diagnostic rather than as evidence of
better preference modelling.
Implicit-reward accuracy uses the DPO implicit reward
$\Delta_\theta(y)=\log\pi_\theta(y\mid x)-\log\pi_{\mathrm{ref}}(y\mid x)$ with
the frozen base model as $\pi_{\mathrm{ref}}$, applied identically to every
system and \emph{not} optimized by \gemmm{}.
We also report A/B choice accuracy.
As an independent check we re-score held-out generations with Skywork-VL
Reward~\citep{wang2025skyworkvl}. 
Appendix~\ref{app:extrm} repeats the analysis with a second judge.
Behavioural suites are AMBER~\citep{wang2023amber} (CHAIR, Hal, Cover, Cog) and
HallusionBench~\citep{guan2024hallusionbench}. POPE~\citep{li2023pope},
MMHal-Bench~\citep{sun2024aligning}, a safety split, and format/length statistics
can be found in Appendix~\ref{app:full}.
Preference metrics use $n{=}1000$ held-out pairs with Wilson $95\%$ CIs and exact
two-sided McNemar tests; AMBER and HallusionBench use their official scorers.
Hyperparameters can be found in Appendix~\ref{app:hparams} and metric definitions in
Appendix~\ref{app:sources}.

\section{Main results}
\label{sec:results}

\paragraph{Preference alignment.}
In Table~\ref{tab:main}, \gemmm{} is best on entropy-depth preference ($61.7\%$),
objective-agnostic implicit-reward accuracy ($41.4\%$), and A/B choice
($68.3\%$; McNemar $p\le0.009$ vs.\ every baseline).

\begin{table}[!ht]
\caption{MM-RLHF held-out preference evaluation ($n{=}1000$ pairs, IDs
3001--4000). Depth Pref is a diagnostic that shares functional form with the
\gemmm{} training reward (Section~\ref{sec:setup}); implicit reward and A/B are
objective-agnostic. Wilson $95\%$ CIs for implicit reward are in the text.
\best{Bold} = best; \secbest{underline} = second and within $1$ point of best.}
\label{tab:main}
\centering
\setlength{\tabcolsep}{3.2pt}
\small
\begin{tabular}{lccc}
\toprule
Method & Depth Pref\,\% & Implicit rew.\,\% & A/B\,\% \\
\midrule
SFT & 57.9 & 40.9 & 66.4 \\
MM-DPO & 58.1 & \secbest{41.0} & 66.4 \\
mDPO & 58.4 & 40.2 & 66.1 \\
SIMA & 57.3 & 40.8 & 66.0 \\
\midrule
\gemmm{} & \best{61.7} & \best{41.4} & \best{68.3} \\
\bottomrule
\end{tabular}

\vspace{2pt}
{\footnotesize Implicit-reward CIs (Wilson $95\%$): SFT $[37.9,44.0]$,
MM-DPO $[38.0,44.1]$, mDPO $[37.2,43.3]$, SIMA $[37.8,43.9]$,
\gemmm{} $[38.4,44.5]$.}
\end{table}
\FloatBarrier

\paragraph{External judge and behavioural benchmarks.}
In Table~\ref{tab:extrm}, Skywork-VL Reward, which we did not train, ranks
\gemmm{} first ($13.82$); the margin is significant against MM-DPO and SIMA
($p=0.04$). A second judge places the same five systems within noise
(Appendix~\ref{app:extrm}).
Table~\ref{tab:amber}: \gemmm{} has the lowest AMBER hallucination ($20.6$) and
CHAIR ($4.6$) at matched coverage ($56.5$ vs.\ $56.4$--$57.0$).
Table~\ref{tab:hallusion}: \gemmm{} is best on all three HallusionBench
granularities (aAcc $69.82$, qAcc $46.59$, fAcc $45.95$).

\begin{table}[!ht]
\caption{Held-out generations ($n{=}1000$, IDs 3001--4000) scored by Skywork-VL Reward~\citep{wang2025skyworkvl}. 
Because raw reward scores are uncalibrated, we focus on relative comparisons. 
$\Delta$ denotes the paired mean difference (\emph{\gemmm{} minus baseline}) alongside two-sided paired $t$-test $p$-values (\best{bold} indicates $p < 0.05$).
A second reward model is reported in Appendix~\ref{app:extrm}.}
\label{tab:extrm}
\centering
\setlength{\tabcolsep}{4pt}
\small
\begin{tabular}{lccc}
\toprule
Method & Mean rew.\,\metup & Bootstrap $95\%$ CI & $\Delta$\,($p$) \\
\midrule
SFT      & 13.06 & $[11.93,14.23]$ & $+0.76$ (0.11) \\
MM-DPO   & 12.85 & $[11.74,13.96]$ & $+0.97$ (\best{0.04}) \\
mDPO     & \secbest{13.27} & $[12.16,14.43]$ & $+0.55$ (0.24) \\
SIMA     & 12.89 & $[11.76,14.07]$ & $+0.93$ (\best{0.04}) \\
\midrule
\gemmm{} & \best{13.82} & $[12.69,14.98]$ & --- \\
\bottomrule
\end{tabular}
\end{table}

\begin{table}[!ht]
\caption{AMBER~\citep{wang2023amber} generative split. Cover is the control:
lower hallucination is only meaningful at comparable coverage.
\best{Bold} = best; \secbest{underline} = second and within $1$ point of best.}
\label{tab:amber}
\centering
\setlength{\tabcolsep}{5pt}
\small
\begin{tabular}{lcccc}
\toprule
Method & CHAIR\,\metdown & Hal\,\metdown & Cover\,\metup & Cog\,\metdown \\
\midrule
SFT & 5.1 & 24.0 & \best{57.0} & \best{0.9} \\
MM-DPO & 5.0 & 23.4 & \secbest{56.7} & 1.0 \\
mDPO & 5.0 & 23.4 & 56.5 & 1.0 \\
SIMA & 5.1 & 23.9 & 56.4 & 1.0 \\
\midrule
\gemmm{} & \best{4.6} & \best{20.6} & 56.5 & \best{0.9} \\
\bottomrule
\end{tabular}
\end{table}

\begin{table}[!ht]
\caption{HallusionBench~\citep{guan2024hallusionbench}.
aAcc/qAcc/fAcc = all-sample, question-group, figure-level accuracy.
\gemmm{} is best on all three granularities.}
\label{tab:hallusion}
\centering
\setlength{\tabcolsep}{6pt}
\small
\begin{tabular}{lccc}
\toprule
Method & aAcc\,\metup & qAcc\,\metup & fAcc\,\metup \\
\midrule
SFT & 68.66 & 43.96 & 43.35 \\
MM-DPO & 68.66 & 42.86 & 42.20 \\
mDPO & \secbest{69.30} & 44.40 & 43.64 \\
SIMA & 68.56 & 43.74 & 43.93 \\
\midrule
\gemmm{} & \best{69.82} & \best{46.59} & \best{45.95} \\
\bottomrule
\end{tabular}
\end{table}
\FloatBarrier

\section{Analysis}
\label{sec:analysis}

\subsection{An operating frontier, not a single optimum}
The reward in Eq.~\eqref{eq:reward} is governed by two hyperparameters with clear roles:
$\lambda$ balances fork exploration against answer decisiveness, while $\rho$ controls how sharply the reward focuses on the highest-entropy minority of reasoning tokens.
Sweeping these parameters does not yield a single dominant model. Instead, it traces a trade-off frontier across qualitatively distinct behaviors (see Appendix~\ref{app:sweep} for the full sweep). 
Our primary configuration ($\lambda{=}2.0$, $\rho{=}0.05$) maximizes entropy depth and minimizes generative hallucination.
By contrast, a more conservative setting ($\lambda{=}2.5$, $\tau{=}0.8$) produces the most \emph{human-like} outputs—boosting the near-chosen rate to $62.4\%$ (up from $48.9\%$), the beat-rejected rate to $64.3\%$, and average ANLS similarity to the preferred reference to $0.818$, while preserving answer-format compliance at $99.3\%$. Meanwhile, a higher-exploration setting ($\lambda{=}3.0$, $\rho{=}0.05$) delivers our strongest results on public multimodal benchmarks: it reaches $71.71$ aAcc, $49.23$ qAcc, and $50.58$ fAcc on HallusionBench—outperforming every baseline in Table~\ref{tab:hallusion} as well as our primary model—and lowers the MMHal-Bench hallucination rate to $0.260$ (vs.\ $0.292$ for the base model). 
Crucially, evaluating along this frontier also clarifies which operating point to trust. Both alternative configurations perform significantly \emph{worse} on our objective-agnostic implicit-reward metric ($p<0.001$ under a paired test), indicating that their gains on reference-similarity and public benchmarks do not reflect stronger underlying preference modeling. For this reason, we anchor our main claims on the primary configuration rather than cherry-picking the settings with the most flattering surface-level proxies.

\subsection{The entropy-depth metric tracks style as well as substance}
Because the depth diagnostic is an average of entropies, it is sensitive to
response length and formatting.
Across our full sweep of full fine-tuned checkpoints, depth preference correlates
\emph{negatively} with verbosity (Pearson $r=-0.60$ against mean response tokens),
and the primary \gemmm{} run is both the highest-depth and one of the terser
models ($69.3$ mean tokens vs.\ $113.3$ for base).
The same run's answer-format compliance is $66.5\%$, below SFT's $72.4\%$ and well
below the base model's $99.0\%$ (Table~\ref{tab:appgen}); the conservative
operating point above shows that this is a property of the operating point rather
than of the objective.
We report this because it bounds the interpretation of the entropy-depth
diagnostic: part of the depth
gain is a change in entropy \emph{style}, which is precisely why the paper's
behavioural claims rest on AMBER and HallusionBench, and its preference claim on
the objective-agnostic implicit reward.

\subsection{Continued training past the fair budget collapses}
It is tempting to continue training a promising checkpoint, and the
generation-side proxies encourage it: continuing an aggressive recipe from a strong
mid-budget checkpoint produced beat-rejected $92.3\%$ and near-chosen $94.0\%$,
far above anything in Table~\ref{tab:main}.
The outputs were degenerate. Answer-format compliance fell to $0.2\%$, repeated
trigrams appeared in $15.9\%$ of generations, and AMBER hallucination rose to
$82.6$ with CHAIR $31.1$ and Cog $13.8$: the model had learned to emit
high-entropy text that scores well on similarity proxies while abandoning the task
format. A quality gate that inspects format rate, repetition, and sentence loops
before any downstream evaluation caught the collapse, and we recommend such a gate
as standard practice for entropy-based rewards; the entropy-collapse literature
in text RL reports structurally similar dynamics~\citep{cui2025entropy}.
Appendix~\ref{app:phase5} lists the continue-training runs and the excluded
mid-budget checkpoint.

\section{Ablations}
\label{sec:ablations}

To isolate which part of the objective is responsible for the behaviour above,
we vary one component at a time, holding data, budget, initialization,
template, and optimizer fixed, so any difference is due to the reward or the
advantage estimator.
Table~\ref{tab:ablate} reports four variants on the same held-out suite:
no fork term (final-entropy penalty only), fork term only, raw REINFORCE in
place of the group-normalized advantage of Eq.~\eqref{eq:adv}, and a binary
exact-match reward in place of the entropy functional.

\begin{table}
\caption{\gemmm{} ablations on the locked 3000-prompt budget. Every row shares
data, initialization, template, and optimizer; only the reward or the advantage
estimator changes.
\best{Bold} = best; \secbest{underline} = second and within $1$ point of best.}
\label{tab:ablate}
\centering
\setlength{\tabcolsep}{2.2pt}
\small
\begin{tabular}{lcccccc}
\toprule
 & Depth & Impl. & A/B & CHAIR & Hal & Sky. \\
Variant & Pref\,\% & \% & \% & \metdown & \metdown & \metup \\
\midrule
Full \gemmm{} & \best{61.7} & \best{41.4} & \best{68.3} & \best{4.6} & \best{20.6} & \best{13.8} \\
No fork term & 60.8 & 39.9 & 67.0 & 5.4 & 24.5 & \secbest{13.7} \\
Fork term only & 57.8 & 40.9 & 66.9 & 4.8 & 21.7 & 13.1 \\
No group adv. & 55.1 & 28.5 & 64.8 & 5.1 & 23.2 & $-3.8$ \\
Exact-match rew. & 55.3 & 32.9 & 64.5 & 4.9 & 21.2 & 4.3 \\
\bottomrule
\end{tabular}

\vspace{2pt}
{\footnotesize Sky.\ is the mean Skywork-VL Reward of
Table~\ref{tab:extrm}. Paired $\Delta$ vs.\ full method:
no fork $+0.1$ ($p=0.80$), fork only $+0.7$ ($p=0.10$), no group
$+17.6$ ($p<0.001$), exact match $+9.5$ ($p<0.001$).}
\end{table}

Group normalization and the entropy functional are essential: removing either
drops implicit reward far below the SFT control ($40.9$).
Dropping the fork term barely moves preference scores, so on-policy training
alone is not what moves hallucination. The effect is confined to AMBER Hal
($24.5$, back in the baseline cluster).
Fork-only reward recovers part of the grounding benefit but none of the
preference benefit, which is why both terms of Eq.~\eqref{eq:reward} are kept.
Appendix~\ref{app:ablate} reports paired tests and both external judges.

\section{Limitations}
\label{sec:limits}

Our study uses one backbone (4B parameters), one primary corpus, and 3000 preference
prompts. Rankings may change at a larger scale, but preference alignment at small-budgets is
precisely what our methodology is based on.
The entropy-depth metric cannot support a claim of better preference modelling; it is reported as a depth
diagnostic, and our objective-agnostic margin over the baselines
(implicit reward, $+0.4$ over the best baseline) is real but not statistically
significant at $n{=}1000$.
The primary operating point degrades answer-format compliance relative to its SFT
initialization ($66.5\%$ vs.\ $72.4\%$), and a conservative operating point that
preserves format gives up entropy depth; a single setting that does both is not
something we can currently offer.
The two external reward models we use agree on our ablations but not on our
baselines: Skywork-VL Reward ranks \gemmm{} first among the systems compared,
whereas IXC-2.5-Reward places all five within noise (Appendix~\ref{app:extrm}).
Neither is a neutral oracle, and IXC-2.5-Reward in particular scores longer
responses higher \emph{within} a fixed system, so its levels partly track
verbosity. We therefore treat external reward scores as a robustness check 
rather than a measurement of alignment quality.
Finally, all judgements here are automatic, we run no human evaluation, and the
safety split in Appendix~\ref{app:full} shows that hallucination gains do not
automatically transfer to refusal behaviour.

\section{Conclusion}

Multimodal preference alignment at limited data budgets saturates
across pair-construction and image-conditioning choices. We name this regime
entropy-blind preference optimization and introduce \gemmm{}, which replaces
pairwise supervision with an on-policy reward that reads the policy's own
fork-versus-final entropy structure.
Under a locked budget \gemmm{} leads on entropy-depth preference and on
objective-agnostic implicit-reward accuracy, and it reduces AMBER hallucination
from $23.4$ to $20.6$ and CHAIR from $5.0$ to $4.6$ at matched coverage while
leading every baseline on all three HallusionBench granularities.
We also mapped the operating frontier of the objective, quantified a style
confound in the entropy diagnostic, and documented a collapse mode that
generation-side proxies fail to detect.

\clearpage
\bibliographystyle{plainnat}
\bibliography{references_final}

\clearpage
\appendix
\section*{Ethics statement}
Multimodal preference data encodes annotator and model biases, and an objective
that rewards decisive final answers can increase confident assertion as well as
grounding. Our safety measurements are limited to one refusal split and show no
improvement from the alignment stage, so deployed models should be audited
independently for safety, fairness, and calibration rather than assumed safe on
the basis of hallucination benchmarks.

\section*{Reproducibility}
Full hyperparameters are in Appendix~\ref{app:hparams}, the complete benchmark
report in Appendix~\ref{app:full}, and the definition and provenance of every
metric in Appendix~\ref{app:sources}; the prompt template and the exact 3000/1000
ID split are fixed across all systems.
Every number behind a main claim comes from a rule-based or log-probability scorer,
so it can be recomputed exactly. The single judged benchmark is MMHal-Bench
(\texttt{gpt-4o-mini} judge); it is used only in the appendix and in one explicitly
marked side remark, never in a headline comparison.
Per-pair outputs of the implicit-reward metric are retained so that confidence
intervals and paired tests can be reproduced.

\section{Ablation results}
\label{app:ablate}

The numbers in Table~\ref{tab:ablate} are discussed below. Each row answers one
question.
Unlike the baseline comparisons, the two largest
ablation effects are statistically unambiguous: removing group normalization or
replacing the entropy reward by exact match both differ from the full method at
$p<0.001$ on the implicit-reward metric.

\emph{Does the fork term matter?} Removing it (final-entropy penalty only) costs
$0.9$ points of depth, $1.5$ points of implicit reward, and, most tellingly, all
of the hallucination gain: CHAIR returns to $5.4$ and Hal to $24.5$, i.e.\ back
into the baseline cluster of Table~\ref{tab:amber}.
The fork reward, not on-policy training per se, is what moves hallucination.

\emph{Is the final-answer penalty redundant?} No. Fork-only reward drops depth
preference to $57.8$, indistinguishable from the pairwise baselines, and leaves
hallucination at $21.7$. Rewarding exploration without demanding a decisive answer
recovers part of the grounding benefit but none of the preference benefit.

\emph{Is group normalization necessary?} It is the single most important component.
Raw REINFORCE on the same reward loses $6.6$ points of depth ($55.1$), $3.5$ points
of A/B accuracy, and, most severely, $12.9$ points of implicit-reward accuracy
($28.5$, far below the SFT control at $40.9$).
Its generations are also the shortest and least format-compliant in the ablation
set ($51.7$ tokens, $50.9\%$ compliance).
Without a within-prompt baseline the update follows the absolute scale of the
entropy reward, which differs across prompts, so the policy drifts on an axis
unrelated to preference quality.

\emph{Does the entropy reward outperform a correctness signal?} A binary exact-match
reward on the same loop reaches only $55.3$ depth, $64.5$ A/B, and $32.9$ implicit
reward. The gain therefore comes from the entropy \emph{shape} of the response
rather than from on-policy reinforcement of verified answers, which at this budget
is too sparse to learn from.
These two variants, the ones that discard either the entropy functional or its
normalization, are also the only ones that fall \emph{far} below the supervised
control on the objective-agnostic metric ($8.0$ and $12.4$ points), whereas the two
reward-shape ablations stay within $1$ point of it.

\emph{Does an external judge agree?} On the two ablations the paper rests on, yes,
and by margins far larger than any between-system difference in
Table~\ref{tab:extrm}: the last column of Table~\ref{tab:ablate} shows that
removing group normalization costs $17.6$ points of Skywork reward and drives the
mean \emph{negative} ($-3.8$, the only configuration in our study for which that
happens), while replacing the entropy reward with exact match costs $9.5$, both at
$p<0.001$. The second reward model of Appendix~\ref{app:extrm} reproduces both
effects at the same significance.
The fork term is the one component neither reward model supports: dropping it is a
wash here ($13.7$ vs.\ $13.8$, $p=0.80$) and slightly favoured by the second judge.
Its evidence is confined to generative hallucination, and we do not claim
otherwise.

\section{Training hyperparameters}
\label{app:hparams}

\paragraph{Shared.}
Backbone Qwen3-VL-4B-Instruct~\citep{qwen2025qwen3vl}; single A100-40GB;
full fine-tuning with gradient checkpointing; pixel budget
$[200704,401408]$; template \texttt{repo\_limit128};
one epoch over the first 3000 MM-RLHF \texttt{short}-modality IDs; evaluation on
IDs 3001--4000 with $\le256$ new tokens.

\paragraph{SFT.}
Learning rate $2{\times}10^{-5}$, one epoch, batch size $1$; this checkpoint
initializes every preference method including \gemmm{}.

\paragraph{Pairwise baselines.}
$\beta=0.1$, learning rate $1{\times}10^{-5}$, one epoch, 3000 pairs.
MM-DPO uses $\beta$-scaling weight $0.1$; mDPO uses image term weight $1.0$;
the SIMA row samples $k{=}2$ candidates per prompt from the SFT policy and
ranks them with a format/length/repetition critic before pairwise training.

\paragraph{\gemmm{}.}
$\lambda{=}2.0$, $\rho{=}0.05$, $k{=}3$, $\tau{=}0.9$, learning rate
$1{\times}10^{-5}$, $T_{\max}{=}128$, entropy split of Eq.~\eqref{eq:split},
3000 cycles, reward mode \emph{entropy}, advantage mode \emph{group-normalized}.
Operating points: conservative
$(\lambda{=}2.5,\rho{=}0.10,\tau{=}0.8)$; aggressive
$(\lambda{=}3.0,\rho{=}0.05,\tau{=}0.9)$.
A fourth sweep point with a $96$-token budget diverged (depth preference $34.3$,
mean length $204$ tokens), which we include as evidence that the token budget and
$\lambda$ interact.

\section{Complete benchmark suite}
\label{app:full}

Table~\ref{tab:appfull} lists every suite we scored for the systems in the main
comparison, including the metrics we do not build claims on.
POPE is scored on the adversarial split; the base-model row is degenerate
(it answers the yes/no probe almost uniformly, giving an uninformative $100$) and
we therefore draw no conclusion from that column.
MMHal-Bench is the one judged benchmark here (96 items, \texttt{gpt-4o-mini} as
judge, integer score $0$--$6$ plus a binary hallucination flag).
Safety is refusal accuracy on the MM-RLHF safety split.

\begin{table}
\caption{Full benchmark report for the main systems. Reported for transparency;
main-paper claims use only AMBER (Table~\ref{tab:amber}) and HallusionBench
(Table~\ref{tab:hallusion}).}
\label{tab:appfull}
\centering
\setlength{\tabcolsep}{2.6pt}
\footnotesize
\begin{tabular}{lccccc}
\toprule
 & POPE & AMBER & MMHal & MMHal & Refusal \\
Method & F1 & disc.\ F1 & avg\,\metup & hal\,\metdown & acc\,\metup \\
\midrule
Base & 100.0$^{\ddagger}$ & 46.1 & 4.45 & 0.29 & 40.0 \\
SFT & 83.1 & 79.8 & 4.27 & 0.33 & 90.3 \\
MM-DPO & 83.1 & 80.6 & 4.49 & 0.27 & 90.5 \\
mDPO & 82.8 & 80.2 & 4.34 & 0.30 & 90.5 \\
SIMA & 82.8 & 80.0 & 4.33 & 0.29 & 90.1 \\
\gemmm{} & 82.1 & 76.8 & 4.29 & 0.31 & 88.9 \\
\bottomrule
\end{tabular}

\vspace{2pt}
{\footnotesize $^{\ddagger}$Degenerate response pattern; not interpretable.}
\end{table}

\begin{table}
\caption{Generation statistics on the held-out prompts. Format compliance is the
fraction of responses that emit the required answer marker. Most of the
compliance regression relative to the base model is introduced by the supervised
stage that initializes all preference methods.}
\label{tab:appgen}
\centering
\setlength{\tabcolsep}{4pt}
\footnotesize
\begin{tabular}{lccc}
\toprule
Method & Mean tokens & Format\,\% & ANLS \\
\midrule
Base & 113.3 & 99.0 & 0.816 \\
SFT & 69.4 & 72.4 & 0.595 \\
MM-DPO & 70.0 & 72.7 & 0.606 \\
mDPO & 69.7 & 72.1 & 0.605 \\
SIMA & 69.6 & 72.6 & 0.597 \\
\gemmm{} (primary) & 69.3 & 66.5 & 0.638 \\
\gemmm{} (conservative) & 106.4 & 99.3 & 0.818 \\
\bottomrule
\end{tabular}
\end{table}

\section{External reward-model scoring}
\label{app:extrm}

Section~\ref{sec:results} scores held-out generations with Skywork-VL
Reward~\citep{wang2025skyworkvl}. Here we repeat the entire analysis with a second,
architecturally unrelated public reward model,
IXC-2.5-Reward~\citep{zang2025ixcreward}, a scoring head on
InternLM-XComposer-2.5.

\paragraph{Protocol.} Both models are run locally at their released weights
(\texttt{Skywork/Skywork-VL-Reward-7B} and
\texttt{internlm/internlm-xcomposer2d5-7b-reward}) in \texttt{bfloat16}, and are
shown the image and the user turn exactly as presented to the policy. Every system
is scored on the identical 1000 held-out items (IDs 3001--4000) with zero scoring
failures, so all comparisons are paired. Skywork exposes a scalar value head, which
we read directly; IXC exposes a scoring head over its chat template. Neither scale
is calibrated and the two are not comparable to each other, so
Table~\ref{tab:extrm-ixc} should only ever be read within its own column.

\paragraph{Baselines: the second judge sees a tie.}
Under IXC-2.5-Reward the five systems are statistically indistinguishable. Every
paired difference against \gemmm{} is at most $0.03$ on a scale whose bootstrap
intervals span roughly $0.16$, with $p=0.32$--$0.88$
(Table~\ref{tab:extrm-ixc}, upper block). This judge therefore does not reproduce
the Skywork ranking of Table~\ref{tab:extrm}; it also does not contradict it, since
\gemmm{} is not significantly \emph{below} any baseline either. We report the tie
rather than the ranking alone.

\paragraph{Ablations: the second judge agrees.}
On the ablations the two models coincide (Table~\ref{tab:extrm-ixc}, lower block).
Removing the group-normalized advantage costs $0.96$ IXC points and replacing the
entropy reward with exact match costs $0.48$, both at $p<0.001$, matching the
Skywork margins of $17.6$ and $9.5$ points reported in Table~\ref{tab:ablate}.
Both judges also decline to support the fork term: under IXC, dropping it is
significantly \emph{better} than the full method ($1.191$ vs.\ $1.008$,
$p<0.001$), and under Skywork it is a wash ($p=0.80$).

\paragraph{Length sensitivity.}
Within a fixed system, IXC reward is positively rank-correlated with response
length (Spearman $\rho={+}0.35$ for SFT and ${+}0.42$ for \gemmm{}), whereas
Skywork is flat to mildly negative ($\rho={+}0.04$ and $-0.29$). Mean response
length is $69$--$75$ tokens for every system in Table~\ref{tab:extrm-ixc}
(Table~\ref{tab:appgen}), so this does not confound the comparison, but it does
mean IXC levels should not be read as a grounding measurement.

\paragraph{Judge disagreement.}
The two models are not interchangeable. On a vanilla text-only DPO run trained on
the same pairs, which is not one of the objectives compared in this paper, Skywork
assigns the second-highest mean reward we measured while IXC ranks it below every
other tuned system, despite that run producing the shortest and least
format-compliant outputs in our study. Rankings from any single multimodal reward
model should be treated as fragile, which is why we report two.

\begin{table}
\caption{The Section~\ref{sec:results} systems and the Section~\ref{sec:ablations}
ablations re-scored by IXC-2.5-Reward~\citep{zang2025ixcreward} ($n{=}1000$ paired
items). $\Delta$ is the paired mean difference \emph{\gemmm{} minus row}; positive
means \gemmm{} scores higher. This scale is unrelated to the Skywork scale of
Table~\ref{tab:extrm}.}
\label{tab:extrm-ixc}
\centering
\setlength{\tabcolsep}{4pt}
\small
\begin{tabular}{lcc}
\toprule
Method / variant & Mean rew.\,\metup & $\Delta$\,($p$) \\
\midrule
SFT      & 1.026 & $-0.02$ (0.60) \\
MM-DPO   & 1.039 & $-0.03$ (0.36) \\
mDPO     & 1.041 & $-0.03$ (0.32) \\
SIMA     & 1.003 & $+0.01$ (0.88) \\
\gemmm{} & 1.008 & --- \\
\midrule
No fork term     & 1.191 & $-0.18$ ($<$0.001) \\
Fork term only   & 1.067 & $-0.06$ (0.09) \\
No group adv.    & 0.047 & $+0.96$ ($<$0.001) \\
Exact-match rew. & 0.525 & $+0.48$ ($<$0.001) \\
\bottomrule
\end{tabular}

\vspace{2pt}
{\footnotesize Paired bootstrap $95\%$ intervals: SFT $[0.949,1.106]$,
MM-DPO $[0.961,1.116]$, mDPO $[0.962,1.120]$, SIMA $[0.927,1.081]$,
\gemmm{} $[0.922,1.092]$, no fork $[1.108,1.270]$, fork only $[0.992,1.144]$,
no group adv.\ $[-0.031,0.127]$, exact match $[0.447,0.604]$.}
\end{table}

\section{Hyperparameter sweep at the locked budget}
\label{app:sweep}

Table~\ref{tab:appsweep} lists the full $(\lambda,\rho,\tau,T_{\max})$ sweep at
the fair 3000-prompt budget. Beat-rejected and near-chosen are generation-side
proxies computed by comparing a sampled response against the annotated pair;
they respond strongly to verbosity and format and should not be read as
preference accuracy.

\begin{table}
\caption{Phase-4 sweep, all full fine-tunes at 3000 prompts.}
\label{tab:appsweep}
\centering
\setlength{\tabcolsep}{3.4pt}
\footnotesize
\begin{tabular}{lcccc}
\toprule
Setting & Depth\,\% & A/B\,\% & Beat-rej\,\% & Near\,\% \\
\midrule
$\lambda2.5,\tau0.8$ & 53.9 & 69.2 & 64.3 & 62.4 \\
$\lambda2.5,\rho0.05$ & 52.1 & 69.0 & 62.6 & 60.8 \\
$\lambda2.5,T_{96},\eta2e{-}6$ & 51.9 & 68.8 & 62.8 & 62.2 \\
$\lambda3.0,\rho0.05$ & 54.1 & 70.0 & 64.1 & 59.4 \\
$\lambda3.0,\tau1.0$ & 45.1 & 69.6 & 59.8 & 66.8 \\
$\lambda3.0,T_{96}$ & 34.3 & 69.4 & 28.5 & 45.8 \\
\midrule
primary $\lambda2.0,\rho0.05$ & 61.7 & 68.3 & 53.2 & 48.9 \\
\bottomrule
\end{tabular}
\end{table}

\section{Continue-training runs and excluded checkpoints}
\label{app:phase5}

We continued training past the locked 3000-cycle budget by adding another 1500
update cycles on the \emph{same} 3000-prompt pool (total budget $4500$).
These runs are \emph{not} fair against the main comparison, so they are excluded
from all main tables; we report them here only as a stress test of the objective.
We start from two parents: the mid-budget checkpoint at 1500 cycles, and the
primary \gemmm{} checkpoint at 3000 cycles (the main-paper model).
Each row changes a small set of knobs relative to that parent
(Table~\ref{tab:appphase5}).
Two continuations from the primary retain its depth score; one continue from the
mid-budget parent collapses as described in Section~\ref{sec:analysis}.
We also exclude the mid-budget parent itself as a headline result: besides the
unfair budget, it degenerates on discriminative suites (POPE F1 $66.7$, near-chance).

\begin{table}
\caption{Continue-training stress tests (+1500 cycles on the same prompts; total
budget 4500). Each row starts from a named parent and changes the listed knobs.
Depth / Beat / Near are held-out preference and generation-side proxies.
Excluded from main tables. Beat/Near above $\sim\!75\%$ co-occur with format
collapse.}
\label{tab:appphase5}
\centering
\setlength{\tabcolsep}{4pt}
\small
\begin{tabular}{cllcccc}
\toprule
\# & Parent checkpoint & Knob changes (+1500 cycles) & Depth & Beat & Near & Note \\
\midrule
1 & Mid-budget @ 1500 cycles & $\lambda{=}2.5$, $\rho{=}0.05$, $\tau{=}0.8$ & 54.5 & 92.3 & 94.0 & collapsed \\
2 & Mid-budget @ 1500 cycles & $\lambda{=}1.5$, lr${=}3{\times}10^{-6}$, $\tau{=}0.7$ & 50.1 & 62.6 & 60.9 & --- \\
3 & Mid-budget @ 1500 cycles & $k{=}3$, $T_{\max}{=}128$, $\lambda{=}2.0$ & 52.8 & 75.3 & 76.6 & --- \\
4 & Primary GEM-MM @ 3000 cycles & $\lambda{=}2.5$, $\tau{=}0.8$, $\rho{=}0.05$ & 62.0 & 44.0 & 45.8 & --- \\
5 & Primary GEM-MM @ 3000 cycles & $\lambda{=}3.0$, $\rho{=}0.05$, lr${=}3{\times}10^{-6}$ & 61.1 & 52.2 & 47.4 & --- \\
\bottomrule
\end{tabular}
\end{table}

\section{Metric definitions and source files}
\label{app:sources}

\emph{Depth preference} counts held-out pairs with
$r(y^{+})>r(y^{-})$ under Eq.~\eqref{eq:reward} using the evaluated model's own
entropies ($\lambda{=}2.0$, $\rho{=}0.05$); it is a diagnostic of optimization
depth, not an objective-agnostic score.
\emph{Implicit-reward preference accuracy} counts pairs with
$\Delta_\theta(y^{+})>\Delta_\theta(y^{-})$ using
$\Delta_\theta(y)=\log\pi_\theta(y\mid x)-\log\pi_{\mathrm{ref}}(y\mid x)$ with the
frozen base policy as reference, from a cached reference log-probability table so
that every system is compared against the identical reference.
\emph{A/B choice accuracy} presents both responses in one prompt and scores the
model's selection.
\emph{Beat-rejected} and \emph{near-chosen} compare a sampled response to the
annotated pair by normalized text similarity and are reported only in appendices.
Per-pair records for the implicit-reward metric are retained, which is what makes
the confidence intervals and McNemar tests possible.


\end{document}